# A Transformer variant for multi-step forecasting of water level and hydrometeorological sensitivity analysis based on explainable artificial intelligence technology

Mingyu Liu[a], Nana Bao[a,b,*], Xingting Yan[c,d*], Chenyang Li[a], Kai Peng[a]

[a] School of Internet, Anhui University, Hefei 230039, P. R. China

[b] National Engineering Research Center for Agro-Ecological Big Data Analysis & Application, Anhui University, Hefei 230039, P. R. China

[c] Forschungszentrum Jülich GmbH, Institute of Energy and Climate Research (IEK), Jülich 52425, Germany

[d] Hefei Institutes of Physical Science, Chinese Academy of Sciences, Hefei 230031, P. R. China

[*]Corresponding author.

E-mail address: x.yan@fz-juelich.de (X. Yan)

**Abstract:**

Understanding the combined influences of meteorological and hydrological factors on water level and flood events is essential, particularly in today's changing climate environments. Transformer, as one kind of the cutting-edge deep learning methods, offers an effective approach to model intricate nonlinear processes, enables the extraction of key features and water level predictions. EXplainable Artificial Intelligence (XAI) methods play important roles in enhancing the understandings of how different factors impact water level. In this study, we propose a Transformer variant by integrating sparse attention mechanism and introducing nonlinear output layer for the decoder module. The variant model is utilized for multi-step forecasting of water level, by considering meteorological and hydrological factors simultaneously. It is shown that the variant model outperforms traditional Transformer across different lead times with respect to various evaluation metrics. The sensitivity analyses based on XAI technology demonstrate the significant influence of meteorological factors on water level evolution, in which temperature is shown to be the most dominant meteorological factor. Therefore, incorporating both meteorological and hydrological factors is necessary for reliable hydrological prediction and flood prevention. In the meantime, XAI technology provides insights into certain predictions, which is beneficial for understanding the prediction results and evaluating the reasonability.

## 1. Introduction

Flood is a dangerous natural disaster that causes extensive devastation, loss of life, destruction of property and damage to public health infrastructure (Jonkman, et al., 2024). Water level is an important indicator of flood, and its fluctuations significantly affect ecological and hydrological systems (Gownaris, et al., 2018). Water level is influenced by numerous factors, including not only hydrological variables, but also meteorological conditions (Quang, et al., 2024). Considering the increasingly drastic climate changes and frequent abnormal climate events nowadays, dynamic processes within the natural cycle, such as rainfall and runoff also change rapidly (Wu, et al., 2023). Therefore, the evolution of water levels usually has the characteristic of large fluctuations, which are highly nonlinear and complex, making the accurate daily water level forecasting difficult (Weng, et al., 2023). Developing a reliable methodology for water level forecasting is crucial for government decision-making and flood risk management (Piadeh, et al., 2023). It is necessary to comprehensively consider multiple meteorological factors and use an accurate model to simulate the nonlinear and complex dynamic process of water level evolution (Han, et al., 2021b; Liao, et al., 2023). Understanding and quantifying the significance of meteorological and hydrological factors in water level fluctuations also play a key role in flood prevention (Lin, et al., 2023; Nikhil Teja, et al., 2023).

Current water level prediction models can be classified into physical and data-driven models (Demir and Yaseen, 2023). The physical models are based on hydrological principles, providing a comprehensive understanding of fundamental forecasting processes (Ozdemir, et al., 2023). However, they face challenges in terms of data collection for numerous physical parameters and accurately simulating the complex nonlinear hydrological behaviors (Guo, et al., 2023). On the other hand, data-driven models, for example statistical (e.g. Auto Regressive, Moving Average, Auto Regressive Moving Average, Auto Regressive Integrated Moving Average, etc.) and machine learning approaches are widely used to address the challenges in nonlinear and non-stationary hydrological time series forecasting (Wang, et al., 2015; Xiang, et al., 2020). The popularity of machine learning models arises from their ability to handle various data and establish correlations among these parameters, eliminating the need for laborious statistical modeling (Sannasi Chakravarthy, et al., 2022; Wee, et al., 2021). For example, the utilization of machine learning methods in lake water level forecasting is supported by numerous references (Zhou, et al., 2020; Zhu, et al., 2020).

Deep learning models (Han, et al., 2021a) based on artificial neural networks (ANNs), have been rapidly developed and gradually applied on water level prediction in recent years (Agarwal, et al., 2022; Bouach, 2024; Wan, et al., 2019; Zhang, et al., 2020). Among these deep learning models, the Long Short-Term Memory model (LSTM) is one typical type and has been recently applied to water level forecasting (Gao, et al., 2020; Giang, et al., 2022; Yokoo, et al., 2022). For example, Hu et al employed ANN and LSTM models to simulate the rainfall-runoff process in Fen River basin of China, with results indicating that LSTM model exhibits superior performance compared to ANN models (Hu, et al., 2018). Xu et al applied Convolutional Neural Networks (CNN)-LSTM neural networks to effectively simulate the water level and flow of Hankou station in China, thereby providing valuable technical supports for optimizing flood control strategies at the Three Gorges Power Station (Xu, et al., 2023b). Recently by leveraging the attention mechanism, Transformer (Vaswani, et al., 2017) demonstrates superior capabilities in capturing long-range dependencies compared to models based on LSTM. Transformer has also been applied in water level prediction work. For instance, Xu et al. firstly used Transformer to explore the water level prediction in Poyang Lake (Xu, et al., 2023a); Wei et al. used Transformer to predict the runoff process in the Yangtze River basin (Wei, et al., 2023); Xu et al. utilized Transformer for water level prediction with fluctuations in Yangtze River and applied deep transfer learning method based on Transformer for flood forecasting with sparse data (Xu, et al., 2023c). However, there still remain unsolved issues in current deep learning method research, such as neglecting other types of input data besides hydrological factors (Shrestha, et al., 2021; Yang, et al., 2023) and the low computational efficiency caused by time complexity of the Transformer algorithm (Keles, et al., 2022). To address these issues, this paper simultaneously utilizes various meteorological and hydrological data as inputs in water level prediction. Additionally, we innovatively propose an enhanced Transformer model that incorporates a nonlinear layer before the linear output layer in traditional Transformer architecture to improve prediction accuracy. Furthermore, we employ a novel attention module called sparse attention mechanism (Zhao, et al., 2019) to enhance the computational efficiency of the self-attention (SA) module.

Advanced machine learning methods demonstrate great capabilities in improving prediction accuracy. However, due to their inherent complexity and opacity, which hinder physical interpretability, it is challenging to understand how these 'black-box' models utilize input variables to make final predictions (Herath, et al., 2021). To address this concern, XAI technologies have recently been proposed to assist in understanding the underlying physical mechanisms involved in AI-based modeling for real-world processes. XAI technologies facilitate gaining insights into these black boxes and elucidating their modeling behaviors (Barredo Arrieta, et al., 2020; Minh, et al., 2022). Understanding the rationale behind specific predictions is as important as model accuracy. Additionally, XAI technology can also be employed to evaluate the contributions of different input features in water level prediction (Hassija, et al., 2024). Therefore, in this work, we utilize an XAI method called SHapley Additive exPlanations (SHAP) (Lundberg and Lee, 2017) for water level predictions. This method helps to illustrate the prediction mechanisms and analyze the sensitivities of meteorological and hydrological factors.

The objective of this study is to develop an enhanced Transformer model that improves both accuracy and time efficiency for water level prediction by simultaneously utilizing various meteorological and hydrological data, as well as quantitatively investigating the dynamics of water levels based on SHAP analysis. To enhance the performance of the Transformer method, an nonlinear output layer and sparse attention mechanism are employed. In this work, meteorological and hydrological data from the Chao Lake basin in China are chosen as the data source. The developed model aims to fully leverage different types of meteorological and hydrological data. By using the SHAP method, novel insights can be obtained regarding the quantitative relationships between water level fluctuations and meteorological as well as hydrological factors. This study is expected to contribute to improving the Transformer algorithm and its application in water level prediction while providing valuable insights into how meteorological-hydrological data influence water levels. The results reported in this work may serve as a reference for effective water resource management and flood prevention in the future.

The rest of this paper is organized as follows: Section 2 presents the data source and methods used in this work, including the model development of the enhanced Transformer, the SHAP method, and model evaluation metrics. Section 3 discusses the main results and corresponding analyses, including model performance analysis and SHAP analysis. Finally, conclusions are given in Section 4.

## 2. Data source and methods
### 2.1 Data source

The Chao Lake basin, one of China's five prominent freshwater lakes, is located in the middle and lower reaches of the Yangtze River. It measures 55 kilometers wide from east to west and covers an area of 780 square kilometers with a lake circumference of 176 kilometers, shown in Fig.1. Both meteorological and hydrological factors contribute to water level fluctuations and floods in this basin.

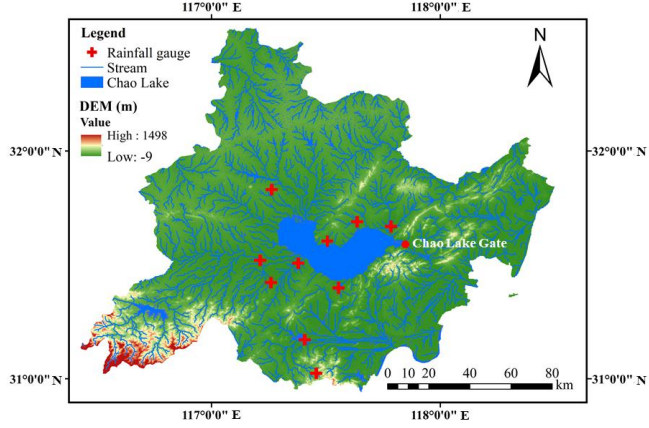

**Fig. 1.** The locations of Chao Lake and the key rainfall stations

Data used for water level prediction for Chao Lake are obtained from two sources: Hydrology-related departments provide rainfall and water level data at the Chao Lake Sluice (http://yc.wswj.net/ahsxx/LOL/), while meteorological data is obtained from the Climate Change Research Centre of the Chinese Academy of Sciences, including day-by-day observations from national state-level stations across China (https://ccrc.iap.ac.cn/). With the raw data, we use bilinear interpolation for downscaling and missing value filling, and specify the study area using vector boundary data of the Chao Lake basin, by utilizing software such as MATLAB, Python, ArcGIS, etc. The dataset covers the time period from 1980 to 2015, the spatial resolution of meteorological data is refined to be $0.05°×0.05°$. The final dataset consists of a total length of 13,148 units. Detailed meteorological and hydrological features used in this work are listed in Table. 1. The dataset is divided into training (70%), testing (20%), and validation sets (10%).

**Table 1.** List of model inputs.

| Feature group | Abbreviation | Description |
| --- | --- | --- |
| Meteorological features | tm | Daily mean temperature (°C) |
|  | pre | Daily precipitation (mm/d) |
|  | tmax | Daily maximum temperature (°C) |
|  | tmin | Daily minimum temperature (°C) |
|  | ssd | Duration of sunshine (h/d) |
|  | win | Daily mean wind speed (m/s) |
|  | rhu | Daily averaged relative humidity (%) |
| Hydrological features | ch_wl | Water level above Chao Lake Gate (m) |
|  | ch_pre | Daily rainfall at Chao Lake Gate (mm) |
|  | qk_pre | Daily rainfall at Quekou Gate (mm) |
|  | zm_pre | Daily rainfall at Zhongmiao Gate (mm) |
|  | ty_pre | Daily rainfall at Tongyang Gate (mm) |
|  | xg_pre | Daily rainfall at Xiage Gate (mm) |
|  | zh_pre | Daily rainfall at Zhaohe Gate (mm) |
|  | zq_pre | Daily rainfall at Zhuanqiao Gate (mm) |
|  | lj_pre | Daily rainfall at Lujiang Gate (mm) |
|  | jn_pre | Daily rainfall at Jinniu Gate (mm) |
|  | nh_pre | Daily rainfall at Nihe Gate (mm) |
|  | tc_pre | Daily rainfall at Tangchi Gate (mm) |

## 2.2 Transformer-based models
### 2.2.1 Transformer-based models

The Transformer demonstrates exceptional global perception and effectively focuses on specific segments of the source sequence through the attention mechanism. As shown in Fig. 2, the Transformer is fundamentally an encoder-decoder model. The data processing procedure can be summarized as follows: Input data undergo embedding and positional encoding operations to retain both numerical and sequential features; Subsequently, the encoder maps these input representations to another continuous representation sequence. Each encoder layer ($N$ layers in total) consists of two sub-layers: a multi-head attention mechanism (MHA) and one position-wise fully connected feed-forward network (FFN). Additionally, residual connections surround each sub-layer, which are then followed by layer normalization. The output from the encoder is further processed by the decoder, taking into consideration previously generated symbols when predicting the final output at each time step. The decoder comprises three sub-layers: Masked multi-head attention mechanism (MMHA), MHA, and FFN. Similar to the encoder, residual connections are employed around each sub-layer and followed by layer normalization (Vaswani, et al., 2017).

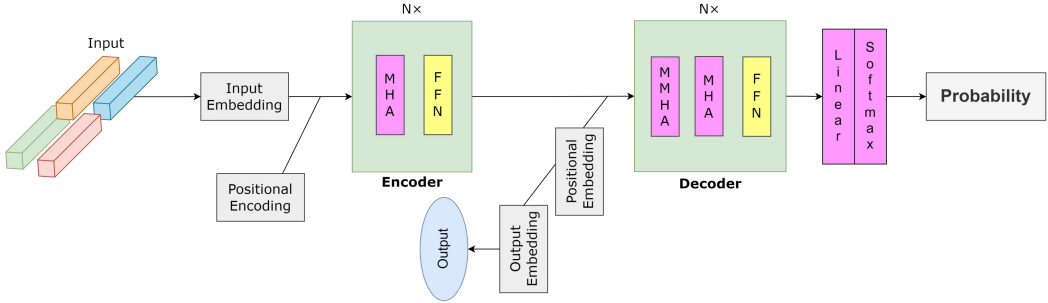

**Fig. 2. Structure of basic Transformer**

The basic Transformer is enhanced on the following two aspects:

(1) Sparse attention

Self-attention, also known as intra-attention, is an attention mechanism that relates different positions within a single sequence to compute a representation of the sequence. The attention function takes a query and a set of key-value pairs as input and produces an output. The input involves linear transformations of the queries, keys, and values, while the output is computed by taking a weighted sum of the values. The value can be determined based on the particular model utilized. All components - queries, keys, values, and outputs - are represented as vectors. In practical applications, we use matrix $Q$ to represent a set of queries with dimension $d_q$. Similarly, matrices $K$ and $V$ are used to represent key vectors with dimension $d_k$ and value vectors with dimension $d_v$ respectively (Vaswani, et al., 2017). Thus, the self-attention output can be calculated as follows:

$$\text{Self attention}(Q, K, V) = \text{softmax}(\frac{QK^{\text{T}}}{\sqrt{d_k}})V \quad (1)$$

Sparse-attention is a modified version of the self-attention model that introduces sparsity in comparison to the traditional approach. Unlike the traditional method of calculating attention weights, sparse-attention employs a masking operation called Mask to sample (as described in Eq. 3) in order to determine the sparsity of vectors. It assigns probabilities exclusively to the most significant elements within the input sequence and subsequently computes attention scores specifically for these selected positions (Zhao, et al., 2019). Similar to self-attention, sparse attention initially calculates attention scores as follows:

$$P = \frac{QK^T}{\sqrt{d_k}} \tag{2}$$

Assuming that higher scores indicate stronger relevance, the Mask (·) operation selects elements with the highest scores and records their positions in a position matrix (*i*, *j*). The sparse-attention masking operation Mask (·) can be written as follows:

$$\text{Mask}(P,k)_{ij} = \begin{cases} P_{ij} & \text{if } P_{ij} \geq t_i \\ -\infty & \text{if } P_{ij} < t_i \end{cases} \tag{3}$$

where $t_i$ represents the *k*-th largest value in the *i*-th row. After performing top-*k* selections, the output of sparse-attention can be calculated using the following formula:

$$\text{Sparse attention}(Q,K,V) = \text{softmax}(\text{Mask}(P,k))V \tag{4}$$

The steps of sparse attention calculation involve implementing a computational process that selectively samples data while disregarding interference caused by irrelevant information. It effectively preserves crucial information while eliminating noise by allowing the model's attention to focus more on elements that contribute stronger relevance. This technique of sparsity significantly reduces the computational complexity and enhances the efficiency of the model. Comparison between self-attention and sparse attention structures is shown in Fig. 3.

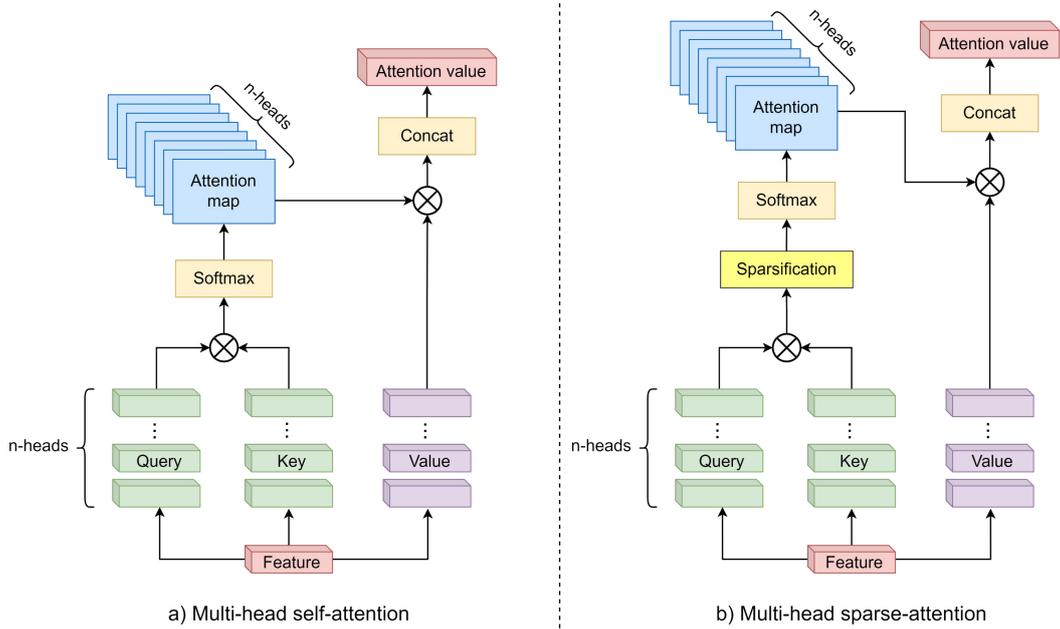

**Fig. 3. Comparison between a) multi-head self-attention and b) multi-head sparse-attention**

Fig. 3 also illustrates the utilization of multi-head self-attention, which enables concurrent computation of output for each head. Specifically, the attention mechanism applies *n* unique and trainable linear transformations to the query, key, and value vectors individually. Subsequently, attention function is performed simultaneously on each transformed projection of query, key, and value to generate *d*-dimensional output values. These outputs are then concatenated and projected once again to obtain the final output. For example,

$$\text{head}_i = \text{Attention}(Q_i, K_i, V_i) \tag{5}$$

where $Q_i, K_i, V_i$ are the mapping of the input obtained from the $i$-th ( $i \in [1, n]$ ) linear projection. Then, all head$_i$ are connected to obtain the output of multi-head attention as follows:

$$\text{MultiHead}(Q, K, V) = \text{Concat}(\text{head}_1, \text{head}_2, ..., \text{head}_n) \tag{6}$$

(2) Nonlinear output layer

In Fig. 4 (a), the basic Transformer is shown in a simplified structure, where the output of decoder is passed to a simple linear mapping to obtain the final output. In this paper, we modify the linear output mapping layer of the basic Transformer model with a nonlinear activation function. Six commonly used nonlinear activation functions are tested, i.e., *ReLU* (Maas, 2013), *Sigmoid* (Nair and Hinton, 2010), *Tanh*, *LeakyReLU* (Racheal, et al., 2023), *ELU* (Clevert, et al., 2015) and *Softplus* (LeCun, et al., 1989; McCulloch and Pitts, 1943). We ultimately select *Tanh* as it performs the best and remains relatively simple when applied in one nonlinear layer. The dashed box intuitively illustrates our modification. This improved structure not only enhances the model's expressive capability but also restricts the output of the first linear layer within (-1, 1) to improve training stability and alleviate gradient vanishing problems. It helps models better adapt to nonlinear and non-stationary relationships in water level data by learning specific patterns within them. Additionally, introducing nonlinear layers can increase the fitting degree to training data.

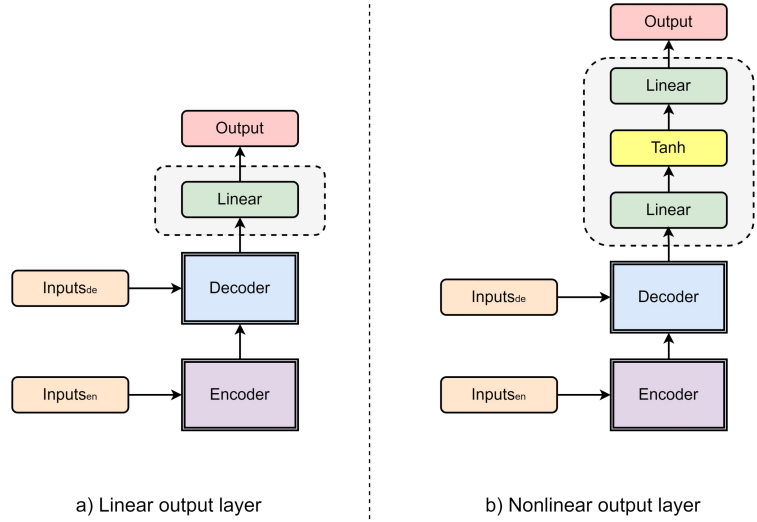

a) Linear output layer     b) Nonlinear output layer

**Fig. 4. Comparison between a) linear output layer and b) nonlinear output layer**

### 2.2.2 Evaluation metrics

The evaluation metrics used in this study include the coefficient of determination ($R^2$), root mean square error (*RMSE*), mean absolute error (*MAE*) and mean bias error (*MBE*). $R^2$ measures the degree of determination between predicted and actual values, *MAE* and *RMSE* measure the average difference between predicted and actual values, *MBE* measures whether the model underestimate or overestimate the prediction target (Castangia, et al., 2023; Ritter and Muñoz-Carpena, 2013). $R^2$, *MAE*, *RMSE*, and *MBE* are calculated as follows:

$$R^2 = 1 - \frac{\sum_{i=1}^{n}(\hat{y}_i - y_i)^2}{\sum_{i=1}^{n}(\hat{y}_i - \overline{y})^2} \tag{9}$$

$$MAE = \frac{\sum_{i=1}^{n}|y_i - \hat{y}_i|}{n} \qquad (10)$$

$$RMSE = \sqrt{\frac{\sum_{i=1}^{n}(y_i - \hat{y}_i)^2}{n}} \qquad (11)$$

$$MBE = \frac{\sum_{i=1}^{n}(y_i - \hat{y}_i)}{n} \qquad (12)$$

The performance of a model can be evaluated by comparing $R^2$, *MAE*, *RMSE*, and *MBE*. Closer value of $R^2$ to 1 indicates better performance, while closer values of *MAE*, *RMSE*, and *MBE* to 0 indicate better performance.

## 2.3 SHAP

SHAP technique has the ability to interpret a prediction model's output. It was originally developed in game theory to determine an individual player's contribution in a collaborative game. The SHAP values provide a solution for fair rewards and assign unique values determined by local accuracy, consistency, and null effect. SHAP provides insights into the model prediction regarding to the significance and positive or negative impact of each input variable. By employing an explanation model with a trained machine learning model and input variables, SHAP can effectively determine the contribution of each variable to the overall model (Lundberg and Lee, 2017). The explanation model can be represented as:

$$E = \phi_0 + \sum_{i=1}^{n}\phi_i t_i \qquad (7)$$

where *E*, *I* and *t* represent the explanation model, the number of input variables and the simplification of the feature vector and $\phi_i \in R$ denotes the contribution of each variable to the machine learning model. The contribution of feature *I*, i.e., $\phi_i$, can be written as:

$$\phi_i(M, x) = \sum_{t \in x} \frac{|t|!(n-|t|-1)!}{n!}[M(t) - M(t \setminus i)] \qquad (8)$$

where *M* and *x* denote the machine learning model and the input variables, and \ is the difference-set notation for set operations.

## 3. Experimental results and discussion

In this section, we evaluate the performance of the developed Transformer variant across different lead times with respect to three Transformer models and the widely used LSTM.

## 3.1 Experimental configurations

Transformer-related and LSTM models are implemented using PyTorch and TensorFlow frameworks. All training and testing of above models are conducted on a Linux environment equipped with a Tesla V100 PCIe 32GB graphics accelerator, utilizing various open-source libraries including torch, TensorFlow, scikit-learn, NumPy, pandas. The optimal set of hyperparameters based on historical experimental results can be determined using the grid search algorithm, where details can refer to Table 2.

Table 2. Hyperparameter settings of models.

| Hyperparameter | | Value |
|---|---|---|
| Transformers | Number of heads | 8 |
| | Number of encoder layers | 1 |
| | Number of decoder layers | 2 |
| | Dimension of fully connected network | 2048 |
| | Dimension of model | 512 |
| | Error function | MSE |
| | Batch size | 32 |
| | Learning rate | 0.0001 |
| | Early stopping patience | 5 |
| | Epoch | 50 |
| | Activation function | Tanh |
| LSTM | Number of layers | 2 |
| | Hidden sizes | 32 |
| | Error function | MSE |
| | Batch size | 32 |
| | Learning rate | 0.0001 |
| | Optimizer | Adam |
| | Early stopping patience | 10 |
| | Epoch | 1000 |

### 3.2 Model Performances

The basic Transformer model demonstrates its high accuracy in predicting water levels. For 1, 3, 5, and 7 days ahead predictions, $R^2$ above 0.789, *MAE* below 0.161, *RMSE* below 0.230, and *MBE* absolute values below 0.036 are obtained using basic Transformer model. Specifically, for 1-day ahead prediction, the $R^2$ is 0.895, *MAE* is 0.108, *RMSE* is 0.159, and *MBE* is 0.025. The results highlight the effectiveness of the Transformer model in water level prediction. However, further improvements can be achieved by incorporating sparse attention mechanism and nonlinear output layer. To evaluate the impact of these enhancements on performance, we compare five models: baseline model (LSTM), Basic Transformer (Transformer), Transformer with sparse attention (Transformer-SPA), Transformer with nonlinear output layer (Transformer-NO), and Transformer variant, enhanced by both sparse attention and nonlinear output layer (Transformer-EN).

Through experimentation, it is observed that all four transformer-based models consistently exhibit a progressive reduction in both training and validation losses throughout the training process. Initially, upon commencement of training, both losses display a rapid decrease, albeit with a gradual slowing down of this trend as epochs increase. Eventually, the losses stabilize at relatively low levels. It is noteworthy that for the Transformer-NO and Transformer-EN models, which incorporate a nonlinear output layer, the validation loss tends to slightly exceed the training loss. However, this disparity is not significant, indicating the effective performance of the nonlinear output layer in capturing complex relationships within the task of water level prediction and successfully avoiding overfitting. Additionally, compared to other models, the variance of the validation loss for the Transformer-EN model is significantly reduced, further demonstrating the positive impact of integrating sparse attention mechanisms and a nonlinear output layer on enhancing model efficiency and generalization capabilities. Further details can be found in Appendix A.

**Table 3.** Evaluation metrics of four transformer-based models and LSTM

| Models | $R^2$ | | | | MAE | | | | RMSE | | | | MBE | | | |
|---|---|---|---|---|---|---|---|---|---|---|---|---|---|---|---|---|
| | 1d | 3d | 5d | 7d | 1d | 3d | 5d | 7d | 1d | 3d | 5d | 7d | 1d | 3d | 5d | 7d |
| LSTM | 0.8848 | 0.8394 | 0.7649 | 0.6831 | 0.1148 | 0.1343 | 0.1571 | 0.1814 | 0.1575 | 0.1858 | 0.2245 | 0.2603 | 0.0734 | 0.0754 | 0.0693 | 0.0626 |
| Transformer | 0.8951 | 0.8679 | 0.8406 | 0.7893 | 0.1083 | 0.1272 | 0.1445 | 0.1607 | 0.1589 | 0.1826 | 0.2027 | 0.2296 | 0.0251 | 0.0314 | 0.0375 | 0.0359 |
| Transformer-NO | 0.9171 | 0.9005 | 0.8777 | 0.8454 | 0.0993 | 0.1104 | 0.1246 | 0.1436 | 0.1467 | 0.1596 | 0.1806 | 0.2059 | 0.0229 | 0.0302 | 0.0324 | 0.0427 |
| Transformer-SPA | 0.9688 | 0.9281 | 0.9024 | 0.8861 | 0.0617 | 0.1034 | 0.1109 | 0.1313 | 0.0967 | 0.1457 | 0.1648 | 0.1941 | -0.0172 | 0.0652 | 0.0193 | **0.0212** |
| Transformer-EN | **0.9762** | **0.9553** | **0.9273** | **0.8891** | **0.0557** | **0.0824** | **0.0971** | **0.1284** | **0.0887** | **0.1273** | **0.1527** | **0.1937** | **-0.0024** | **0.0097** | **-0.0134** | 0.0249 |

The Transformer-EN model has shown superior performance than other models for different lead times in terms of four evaluation metrics. Table 3 provides the values of $R^2$, *MAE*, *RMSE* and *MBE* for each model in the testing set. The performance of the LSTM model is comparable to that of the basic Transformer model for a 1-day lead time. However, as the lead time increases, there is a gradual widening in the performance gap between these two models. Taking $R^2$ as an example, the LSTM model exhibits a reduction in its $R^2$ value by 0.01, 0.03, 0.08, and 0.11 compared to the basic Transformer model across lead times ranging from 1 to 7 days. The Transformer-EN model, our proposed variant, surpasses the basic Transformer in terms of performance. For example, at a 1-day lead time, Transformer-EN enhances the $R^2$ value by 0.08 and concurrently reduces *MAE*, *RMSE*, and *MBE* by 0.05, 0.07, and 0.02 respectively. Similar enhancements are observed for lead times of 3 to 7 days. The performance metrics of five models are clearly shown in Fig. 5. The horizontal axis represents different prediction periods, namely, 1 day, 3 days, 5 days, and 7 days. The vertical axis represents various performance metrics: $R^2$, *MAE*, *RMSE* and *MBE*. The pink, red, green, blue, and yellow boxes represent different models, i.e. LSTM, Transformer, Transformer-NO, Transformer-SPA, and Transformer-EN. Utilizing solely nonlinear output layer (Transformer-NO) and sparse attention mechanism (Transformer-SPA) both improve prediction performance. By simultaneously incorporating nonlinear output layer and sparse attention mechanism, Transformer-EN further achieves the best performance while reducing time complexity as well.

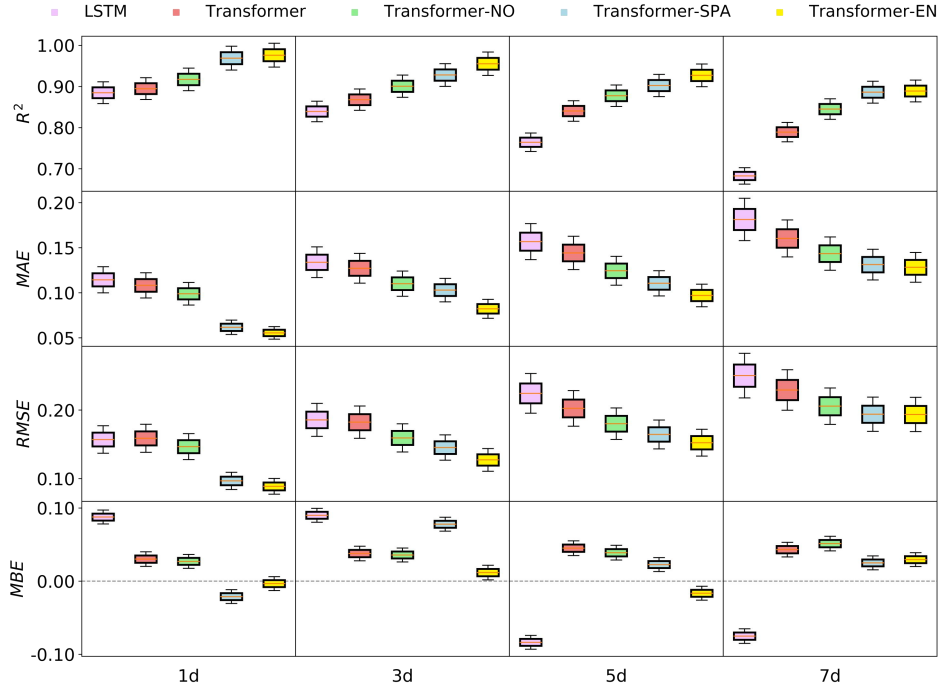

**Fig. 5.** Evaluation metrics of four transformer-based models and LSTM

Fig. 6 shows the prediction results of water level evolution from June 15th to August 15th, 2015, during which frequent fluctuations in water level occurred. Transformer-EN successfully captures the characteristics of water level evolution, especially the typical peaks and valleys (denoted by red rectangles) during this period, exhibiting better predictions to actual values than other three models.

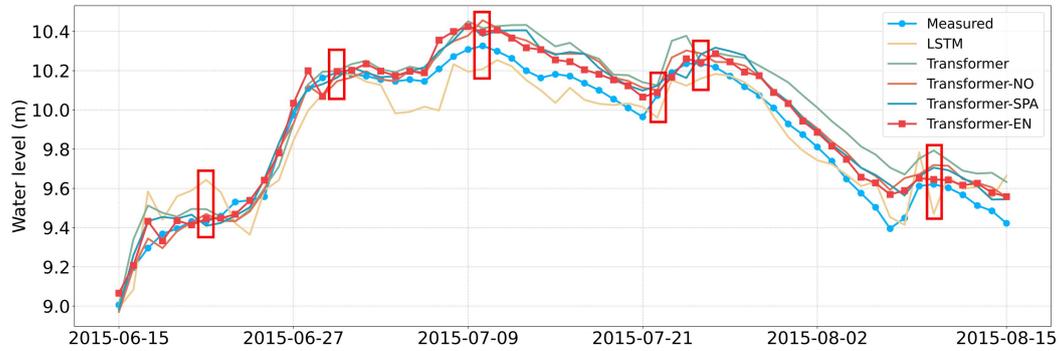

**Fig. 6. The water level prediction performance of four Transformer based models from June 15th to August 15th in 2015 with prediction length of 1 day**

Fig. 7 compares the results for different lead times (from 1 to 7 days), with particular emphasis on water level peaks. Transformer-EN demonstrates good performance in predicting water level across the entire testing set (Fig. 7(a)). However, as the lead time increases, the prediction accuracy of Transformer-EN slightly decreases (Figs. 7(b), (c) and (d)). Nevertheless, even for 7-day lead time, Transformer-EN still maintains a relatively good accuracy. Overall, Transformer-EN exhibits outstanding performance in water level prediction tasks.

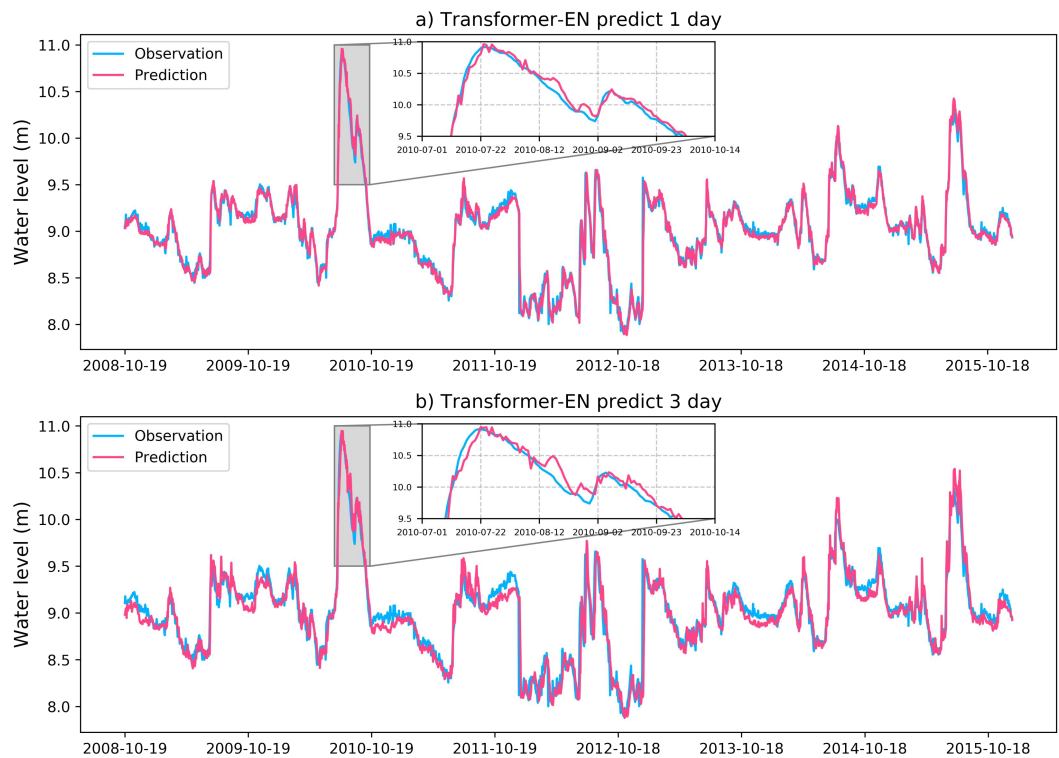

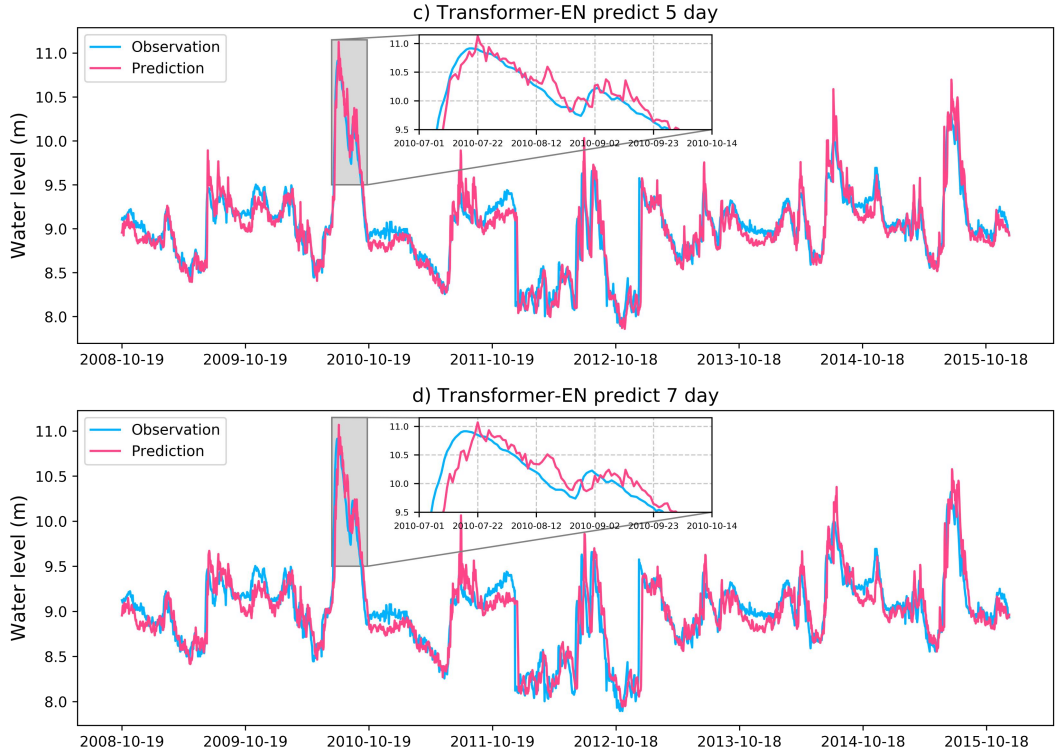

**Fig. 7. Prediction of water level using enhanced Transformer for different lead times**

### 3.3 SHAP analysis

Contributions of different input features to the output are analyzed using SHAP tool, which facilitates the identification of important features for water level prediction. The contributions of 19 different input features in total are compared in Fig. 8. Fig. 8(a) compares the contribution degrees of different input features, and Fig. 8(b) shows the percentage. The hydrological data contributes to the degree of 65.1%, with water level data alone contributing 51.2% and rainfall from other gates contributing 13.9%. It is noteworthy that meteorological data within the Chao Lake basin contributes to the degree of 34.9%, with daily minimum temperature, mean temperature, and maximum temperature contributing to degrees of 10.0%, 9.2%, and 6.2% respectively. Therefore, water level is strongly related to meteorological data, which indicates that it is essential to consider meteorological factors when constructing hydrological models for accurate water level prediction.

The SHAP values for each input feature across all instances are visualized in Fig. 8(c). Each row represents a feature and each point corresponds to a certain instance, with its color indicating the original value of that specific feature. The position of each point is determined by the corresponding SHAP value, while their density along each feature row reflects clustering patterns. From this figure, it is evident that Chao Lake water level (ch_wl) is the most dominant input feature, followed by meteorological variables such as daily minimum temperature (tmin), daily mean temperature (tm), and daily maximum temperature (tmax), which also significantly impact water level prediction. The importance ranking is consistent with results in Fig.8. Moreover, SHAP analysis shows that the influence degree of a certain feature on the output depends on its specific feature value. For instance, Chao Lake water level (ch_wl) and daily minimum temperature (tmin) generally contribute positively (positive SHAP value) when their own values are high (red dots), while contribute negatively (negative SHAP value) when their own values are low (blue dots). On the contrary, daily mean temperature (tm) and daily maximum temperature (tmax) generally contribute positively when their own values are low, while contribute negatively when their own values are high. Such insights provided by SHAP analysis enhance the interpretability and transparency of deep learning models. More process analyses related to the final predicted value for typical samples can refer to Appendix B.

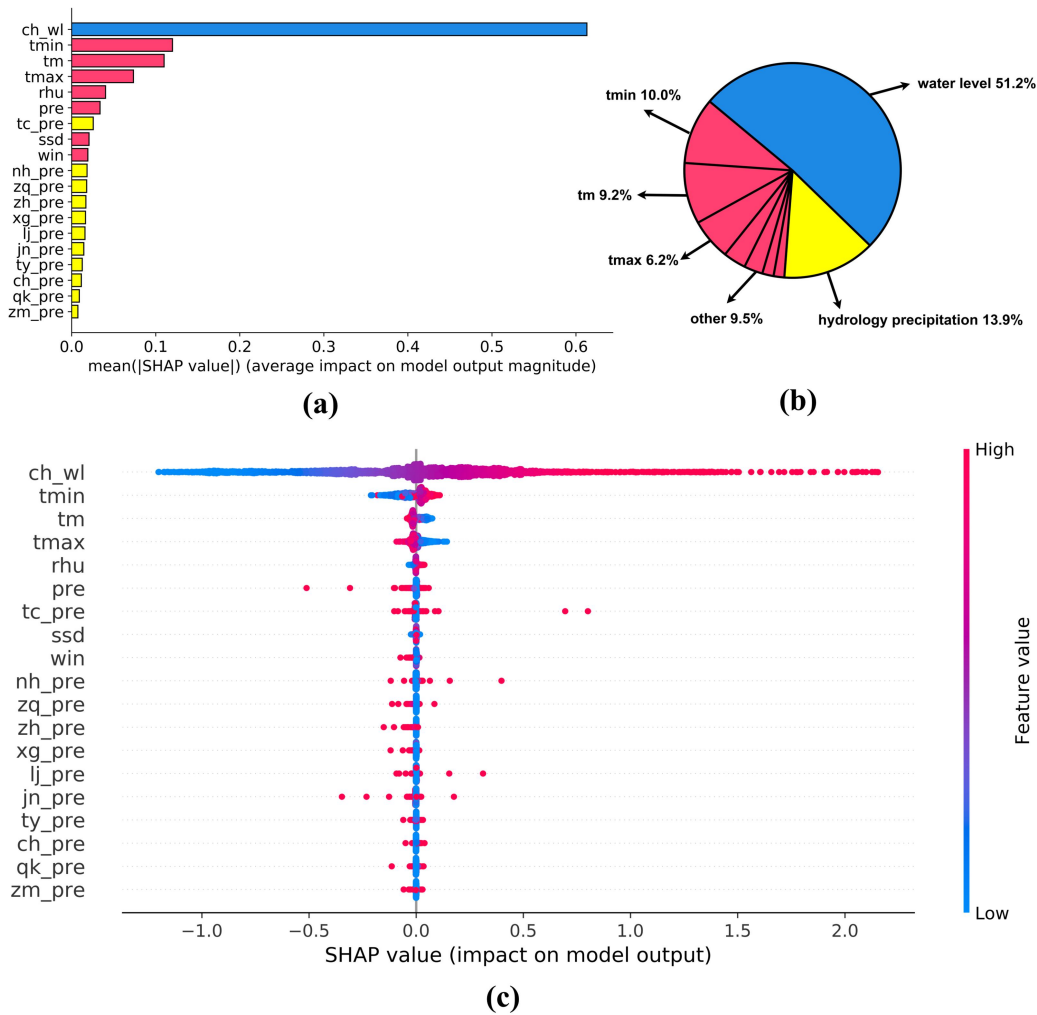

Fig. 8. (a) Contributions of 19 different input features; (b) Percentage of the most important input features; (c) SHAP value for different input features across all data samples.

## 4. Conclusions

In this study, an effective Transformer variant is developed for the analysis of water-level dynamics, and applied in water level prediction by simultaneously utilizing hydrological and meteorological factors. This Transformer variant incorporates sparse attention mechanism and introduces nonlinear output layer in the decoder module, resulting in the best performance compared to other Transformer-based models, with a large $R^2$ of 0.976 and a small *MAE* of 0.056 in 1-day water level prediction, for example. Quantitative analyses of the impacts of different meteorological and hydrological factors are conducted using SHAP XAI technology. The sensitivity analyses reveal the significant impacts of meteorological factors on water level prediction, with a contribution rate as large as 34.9%, only second to the water level itself. In particular, temperatures (tmin, tm, tmax) are shown to be the most important meteorological factors, contributing up to 25.4%. This strongly indicates that water level exhibits strong dependences on meteorological elements, and emphasizes the necessity of simultaneous consideration of both meteorological and hydrological factors in establishing accurate hydrological models, which is significant for improving flood early warning capability. Furthermore, SHAP analyses provide insights into the prediction model, such as whether certain features contribute positively or negatively or how different input features collectively lead to final results in specific predictions etc.

In the future, additional types of meteorological data such as radar, atmospheric pressure, evaporation, etc. can be integrated in the prediction model. The combination of Transformer and Graph Neural Networks will be explored to effectively capture spatial and temporal features. Additionally, diverse XAI methods can be employed to provide deeper insights and interpretabilities into the prediction model.

**Declaration of competing interest**

The authors declare that they have no known competing financial interests or personal relationships that could have appeared to influence the work reported in this paper.

**Acknowledgments**

This work is financially supported by the National Natural Science Foundation of China (62273001). The authors would like to thank the researchers of the Anhui Survey & Design Institute of Water Resources and Hydropower Co., Ltd. and Chaohu Research Institute in China, especially Juan Tian for her helpful discussions and supply of rich hydrological data.

**Appendix A.**

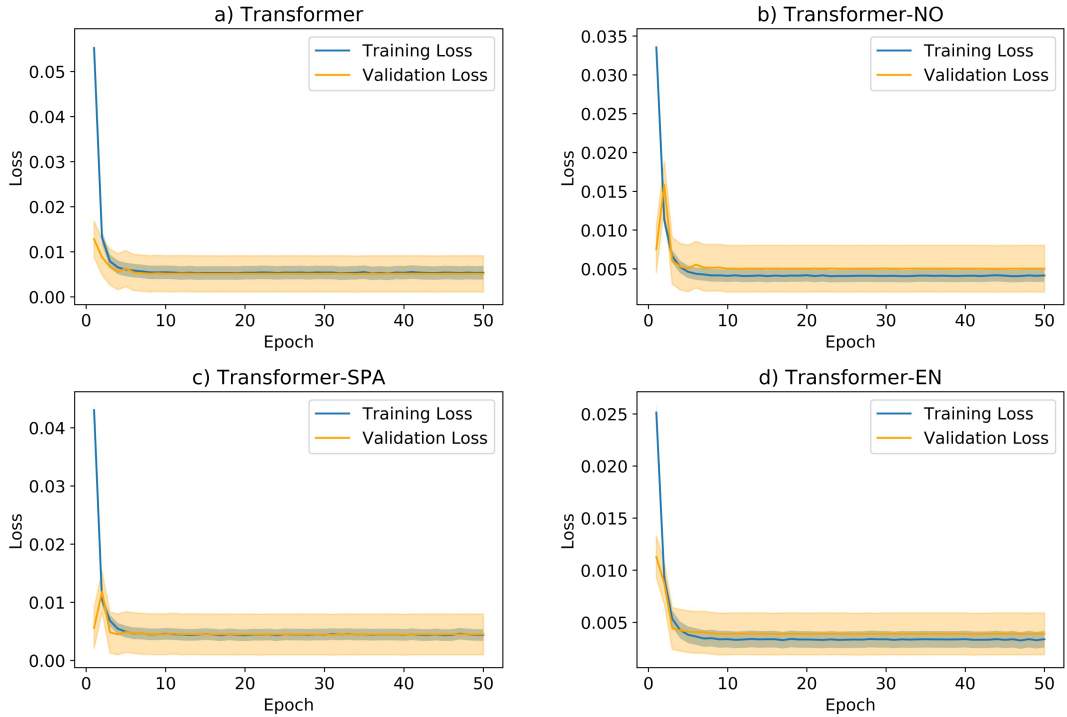

**Fig. 1. Loss vs. Epoch of four transformer-based models.**

As shown in Fig. 1, it is evident that all four transformer-based models consistently exhibit a reduction of training (blue line) and validation (orange line) losses throughout the training process. During the initial stage of training, both training loss and validation loss exhibit rapid decreases. The increase of epoch leads to gradual deceleration of the loss decreasing. At the end of the training process, the loss value saturates at low levels. The validation loss is slightly larger than the training loss for Transformer-NO and Transformer-EN models which have nonlinear output layer, but it still remains at a very small value. This indicates that nonlinear output layer effectively captures complex relationships in water level prediction and achieves good adaptation to new data without overfitting. The variance of validation loss (indicated by orange shaded area) is reduced for Transformer-EN model compared to other models, suggesting that sparse attention and nonlinear output layer contribute enhanced efficiency and generalization.

**Appendix B.**

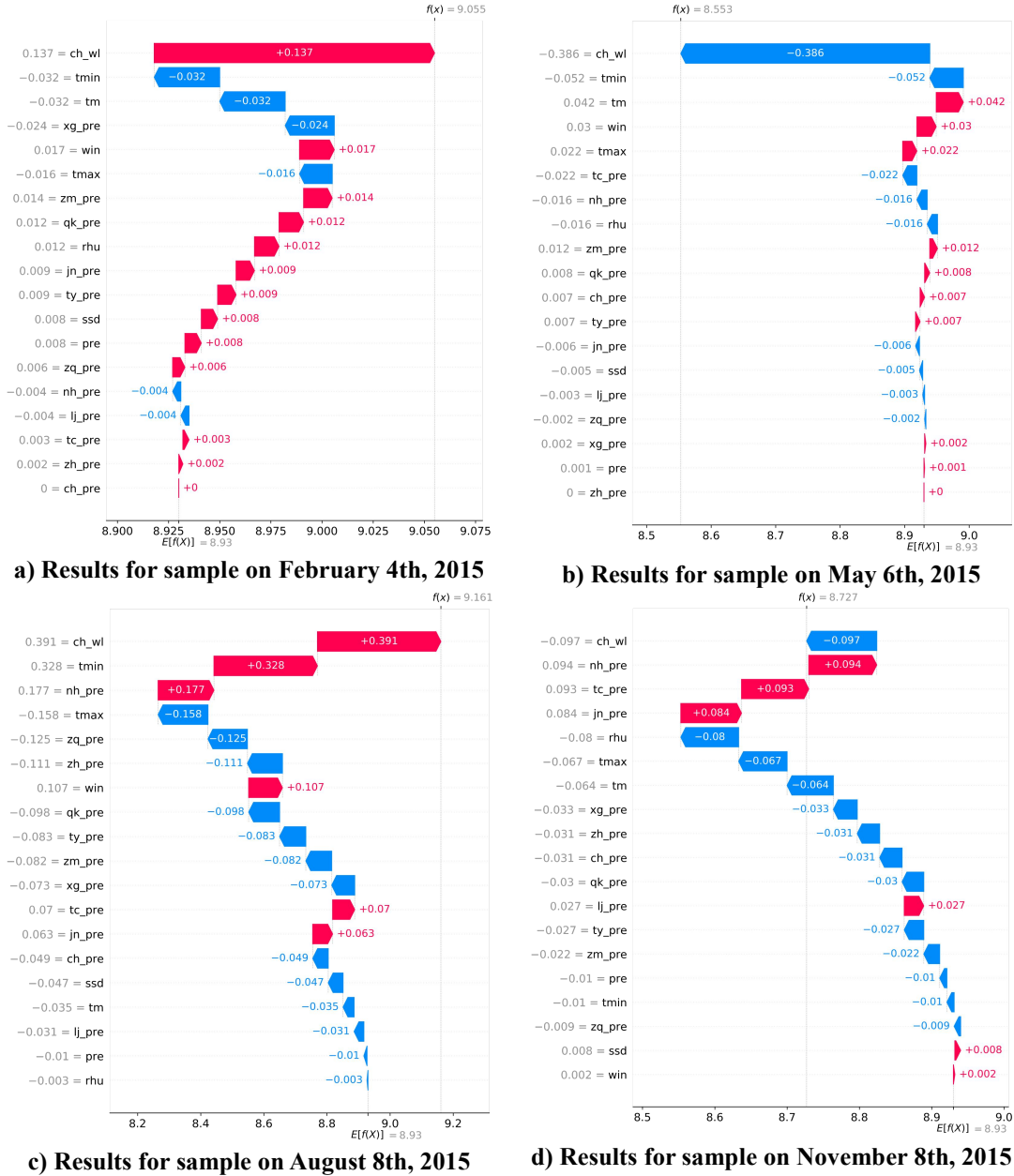

a) Results for sample on February 4th, 2015
b) Results for sample on May 6th, 2015
c) Results for sample on August 8th, 2015
d) Results for sample on November 8th, 2015

Fig. 1. SHAP analyses for specific instances

SHAP analyses for four individual samples can be found in Fig. 1, which illustrates the roles of different input features for certain predictions, i.e. how each feature contributes to the final prediction. These four samples correspond to February 4th, May 6th, August 8th, and November 8th in 2015 in China, roughly representing the beginnings of four seasons with different climate conditions. Red arrows indicate positive contributions while blue arrows indicate negative contributions. The length of arrow reflects the importance of a feature on prediction results. Starting from the average value E[f(x)], each feature exhibits its own contribution, and the final predicted value f(x) is determined by their contributions collectively. The importance ranking varies slightly for different samples. For example, ch_wl and tmin are shown to be the top two most important variables in Fig. 1(a-c), while the factor of nh_pre becomes important in Fig. 1(d).

The SHAP values also vary among different samples. For example, ch_wl positively affects the prediction results in Fig. 1(a) and (c), while negatively affects the prediction results in Fig. 1(b) and (d). This kind of local interpretability provided by SHAP analysis allows us to intuitively understand each feature's role in certain predictions, thereby increasing transparency in forecasting processes and enabling us to clearly identify key features in each prediction along with their respective contributions towards final prediction outcome.